\documentclass[conference]{IEEEtran}
\IEEEoverridecommandlockouts
\usepackage{cite}
\usepackage{amsmath,amssymb,amsfonts}
\usepackage{algorithmic}
\usepackage{graphicx}
\usepackage{textcomp}
\usepackage{xcolor}
\usepackage{subfig}
\usepackage{threeparttable}
\usepackage{booktabs}

\def\BibTeX{{\rm B\kern-.05em{\sc i\kern-.025em b}\kern-.08em
    T\kern-.1667em\lower.7ex\hbox{E}\kern-.125emX}}
\begin{document}

\title{The Matter of Time --- A General and Efficient System for Precise Sensor Synchronization in Robotic Computing}

\author{\IEEEauthorblockN{
Shaoshan Liu\IEEEauthorrefmark{1},
Bo Yu\IEEEauthorrefmark{1},
Yahui Liu\IEEEauthorrefmark{1},
Kunai Zhang\IEEEauthorrefmark{1},
Yisong Qiao\IEEEauthorrefmark{1},
Thomas Yuang Li\IEEEauthorrefmark{1},
Jie Tang\IEEEauthorrefmark{2}\IEEEauthorrefmark{4}\\}
Yuhao Zhu\IEEEauthorrefmark{3},
\IEEEauthorblockA{\IEEEauthorrefmark{1}PerceptIn, U.S.A.\\}
\IEEEauthorblockA{\IEEEauthorrefmark{2}South China University of Technology, China\\}
\IEEEauthorblockA{\IEEEauthorrefmark{3}University of Rochester, U.S.A.\\}
\IEEEauthorblockA{\IEEEauthorrefmark{4}corresponding author cstangjie@scut.edu.cn}}

\maketitle

%!TEX root=paper.tex

\newcommand{\website}[1]{{\tt #1}}
\newcommand{\program}[1]{{\tt #1}}
\newcommand{\benchmark}[1]{{\it #1}}
\newcommand{\fixme}[1]{{\textcolor{red}{\textit{#1}}}}

\newcommand*\circled[2]{\tikz[baseline=(char.base)]{
            \node[shape=circle,fill=black,inner sep=1pt] (char) {\textcolor{#1}{{\footnotesize #2}}};}}

\ifx\figurename\undefined \def\figurename{Figure}\fi
\renewcommand{\figurename}{Fig.}
\renewcommand{\paragraph}[1]{\textbf{#1} }
\newcommand{\figline}{{\vspace*{.05in}\hline}}

\newcommand{\Sect}[1]{Sec.~\ref{#1}}
\newcommand{\Fig}[1]{Fig.~\ref{#1}}
\newcommand{\Tbl}[1]{Tbl.~\ref{#1}}
\newcommand{\Equ}[1]{Equ.~\ref{#1}}
\newcommand{\Apx}[1]{Apdx.~\ref{#1}}
\newcommand{\Alg}[1]{Algo.~\ref{#1}}

\newcommand{\specialcell}[2][c]{\begin{tabular}[#1]{@{}c@{}}#2\end{tabular}}
\newcommand{\note}[1]{\textcolor{red}{#1}}

\newcommand{\proj}{\textsc{Mesorasi}\xspace}
\newcommand{\mode}[1]{\underline{\textsc{#1}}\xspace}
\newcommand{\sys}[1]{\underline{\textsc{#1}}}

\newcommand{\no}[1]{#1}
\renewcommand{\no}[1]{}
\newcommand{\RNum}[1]{\uppercase\expandafter{\romannumeral #1\relax}}

\def\cA{{\mathcal{A}}}
\def\cF{{\mathcal{F}}}
\def\cN{{\mathcal{N}}}

% checkmark and xmark in the pifont package
%\newcommand{\cmark}{\ding{51}}
%\newcommand{\xmark}{\ding{55}}

\begin{abstract}
Time synchronization is a critical task in robotic computing such as autonomous driving. In the past few years, as we developed advanced robotic applications, our synchronization system has evolved as well. In this paper, we first introduce the time synchronization problem and explain the challenges of time synchronization, especially in robotic workloads. Summarizing these challenges, we then present a general hardware synchronization system for robotic computing, which delivers high synchronization accuracy while maintaining low energy and resource consumption. The proposed hardware synchronization system is a key building block in our future robotic products. 
\end{abstract}

\maketitle

\section{Introduction}
\label{sec:intro}

Time synchronization is critical in systems that involve multiple components, such as autonomous vehicles and robots that integrate multiple sensors \cite{liu2020engineering, liu2020creating}. When two sensors, for instance a camera and a LiDAR, take two independent measurements of the robot's environment, the measurements need to happen at the same time in order for the robot to fuse the measurements to reconstruct an accurate and comprehensive view of the environment. Without proper synchronization, multiple sensors could provide inaccurate, ambiguous view of the environment, leading to potentially catastrophic outcomes. 

In this paper, we summarize our past four years' experiences of developing robotic synchronization systems and present a general and efficient sensor synchronization system for autonomous machines such as mobile robots and autonomous vehicles. We believe that this design will help the embedded systems community understand the need for precise synchronization, and help the robotics community to understand how to design a general system for precise synchronization. 

The rest of this paper is as follows. Section \ref{sec:synch} firsts define time synchronization in robotic computing, including intra-machine synchronization and inter-machine synchronization, and  demonstrate why precise synchronization is essential to robotic workloads using real-world data we collected during our daily operations. Section \ref{sec:inext} presents our four principles for designing synchronization systems. Guided by the four principles, Section \ref{sec:hardware} introduces our general hardware synchronization system to address the challenges of both intra-machine and inter-machine time synchronization. While our autonomous vehicles operate on earth, we briefly discuss the challenges of sensor synchronization in space and how our system is expected to be applicable to space scenarios too.

%In our practical experience, in the case of utilizing stereo cameras for depth perception, even if the two cameras are off-sync by only 30 $ms$, the depth estimation error could be as much as 5 $m$. In the case of utilizing IMU and camera for localization, when the IMU and camera are off by 40 $ms$, the localization error could be as much as 10 $m$ \cite{yu2020building}.

%For instance, in the case of cooperative autonomous driving, the systems-on-vehicle (SoVs) needs to synchronize with the systems-on-road  (SoRs) to provide a comprehensive perception of the environment. 

\section{Sensor Synchronization in Robotic Computing}
\label{sec:synch}

\begin{figure}[t]
\centering
\includegraphics[width=1\columnwidth]{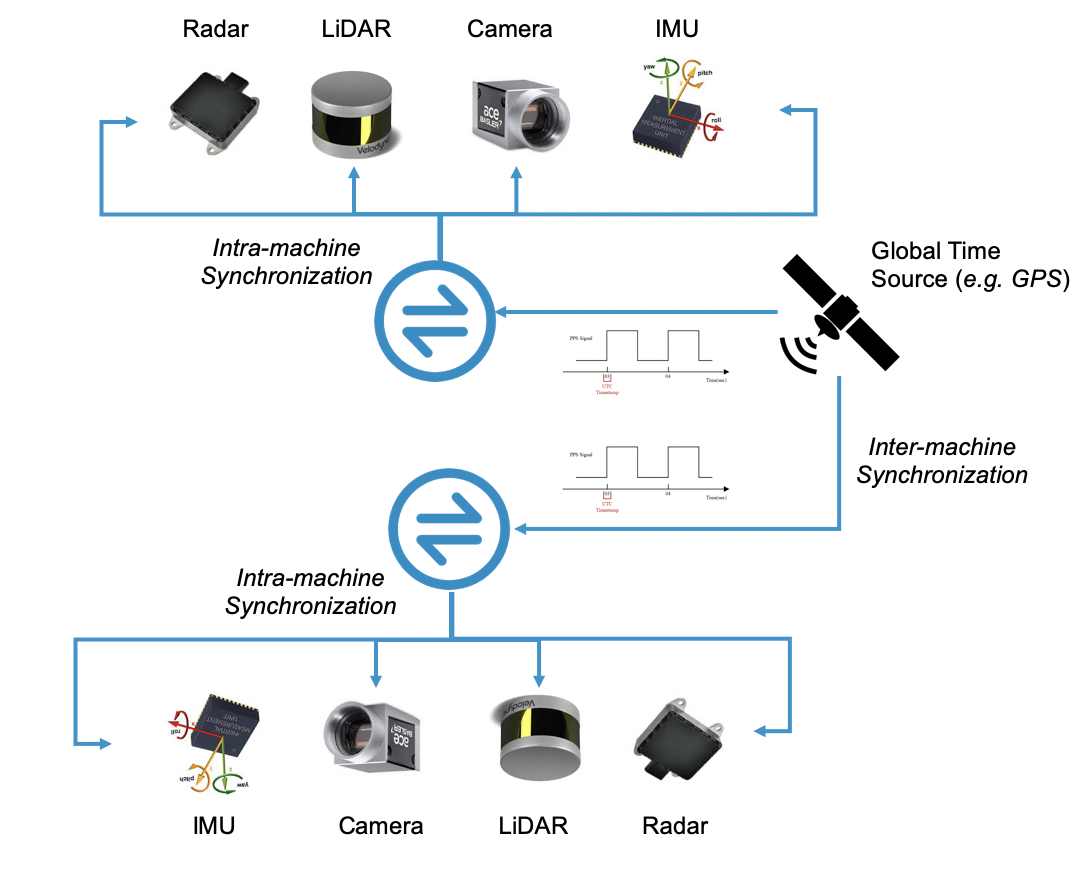}
\caption{Intra-machine and inter-machine synchronizations.}
\label{fig:timeArch}
\vspace{-5mm}
\end{figure} 

The goal of sensor synchronization is to ensure that \textit{sensor samples that have the same timestamp correspond to the same event}. Two kinds of sensor synchronization exist as shown in \Fig{fig:timeArch}. \textbf{Intra-machine synchronization} concerns with synchronizing sensors within a particular autonomous machine. \textbf{Inter-machine synchronization}, instead, concerns with synchronizing sensors across different autonomous machines. Inter-machine synchronization is critical to connected autonomous machines that collaboratively perform a task.
%to associate each sensor sample with a precise timestamp such 

Sensor data in popular datasets, such as EuRoC \cite{burri2016euroc}, are synchronized beforehand to allow researchers to focus on algorithm development and evaluation. Real-world deployment scenarios, however, do not provide well-synchronized data. This section motivates the need for both intra-machine (\Sect{sec:synch:int}) and inter-machine (\Sect{sec:synch:ext}) synchronization using real-world data we collect during our daily operations.

%Fig.~\ref{fig:timeArch} illustrates a general time synchronization architecture for robotic applications. First, each robot can be equipped with multiple sensors, the common ones include LiDARs, cameras, IMUs, and radars, and the key challenge for time synchronization is to have these sensors align to the same time axis with minimum time variation. The main purpose of the \textit{synchronization unit} in the figure is to synchronize all these sensors, which can be done in different layers of the software stack or in hardware.  In addition, for inter-machine synchronization, a global time source is needed.  A common global time source is the pulse-per-second (PPS) signal generated by the atomic clocks on the Global Positioning System (GPS), hence a GPS receiver can be used to receive the PPS signal, and then the \textit{synchronization unit} can align all sensors to the global time source \cite{mogul2000rfc2783}.

\subsection{The Need for Intra-Machine Synchronization}
\label{sec:synch:int}

We use two tasks, perception and localization, to motivate the need for intra-machine synchronization.

\paragraph{Perception} Many perception tasks, such as object detection, increasingly fuse 3D information from LiDARs (i.e., point cloud) and 2D information from cameras~\cite{Geiger2012KITTI}. Thus, LiDARs and cameras must be synchronized.

\begin{figure}[t]
%\vspace{-10pt}
\centering
\subfloat[Three sensor (mis-)synchronization cases.]
{
  \includegraphics[trim=1 1 1 1, clip, width=0.98\columnwidth]{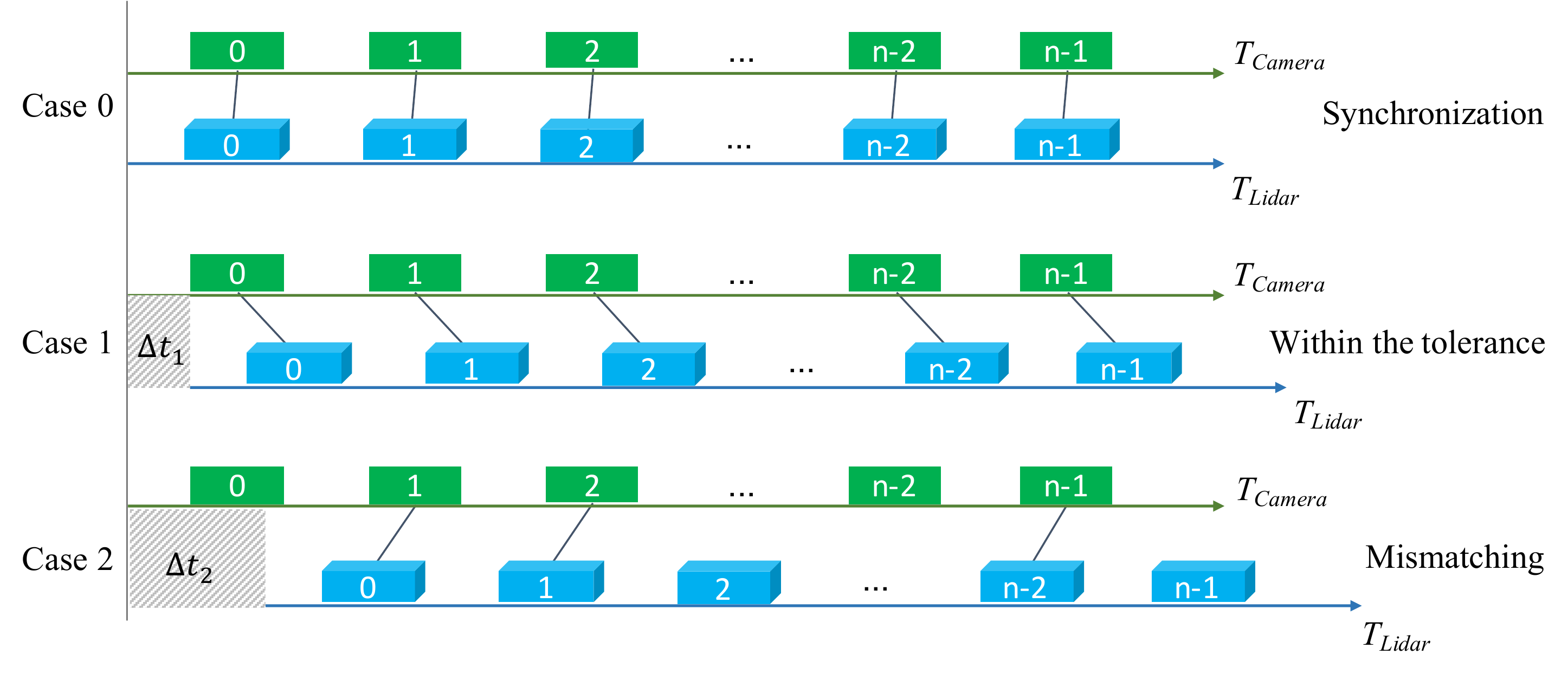}
  \label{fig:axis}
}
\\
%\vspace{10pt}
\subfloat[Object detection using synchronized (left) and mis-synchronized (right) data. The green box denotes a 3D detection bounding box projected onto the 2D image and the orange box denotes a 2D detection bounding box.]
{
  \includegraphics[trim=0 0 0 0, clip, width=0.98\columnwidth]{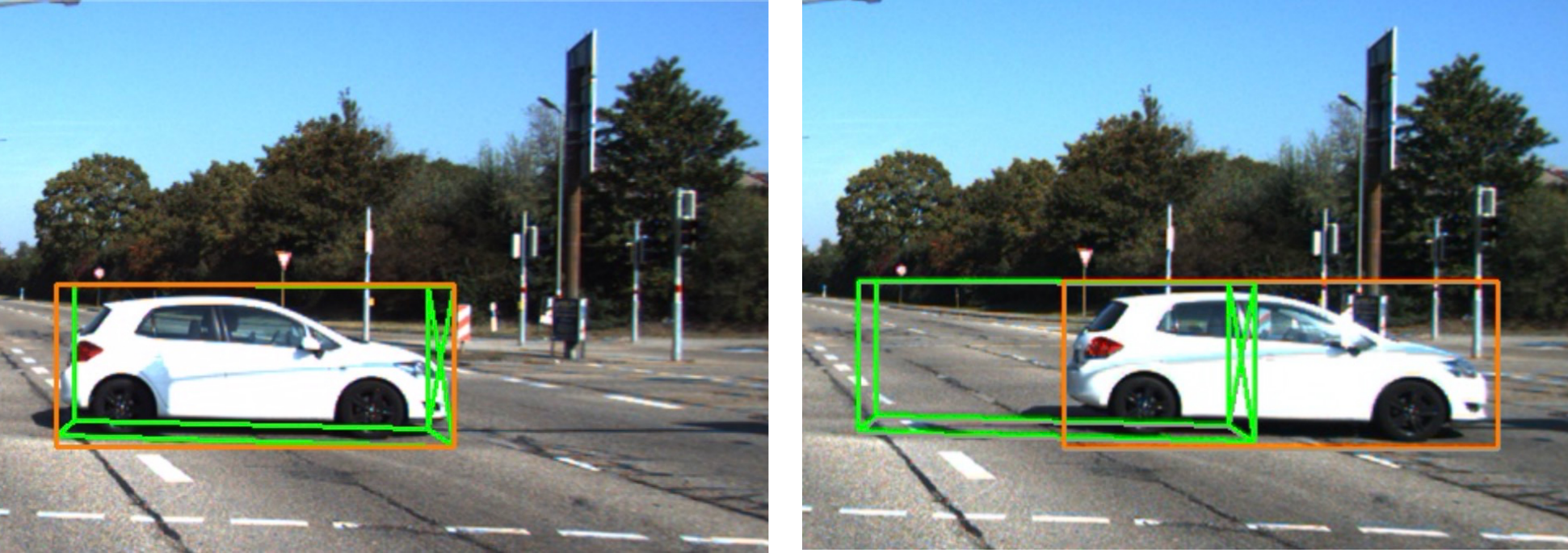}
  \label{fig:bbox}
}
\caption{Impact of intra-machine camera-LiDAR synchronization on object detection.}
\label{fig:box_matching}
\end{figure}

\Fig{fig:axis} presents three time synchronization cases. In the ideal case 0, the LiDAR and the camera are well-synchronized. As a result, \Fig{fig:bbox} (left) shows that the corresponding 3D bounding box detected from LiDAR data and the 2D bounding box detected from camera data are well-matched. In case 1, the triggering time of the camera is ahead of the LiDAR by $\Delta t_1$. If $\Delta t_1$ is within the time tolerance, the 3D and 2D bounding boxes still overlap as shown in \Fig{fig:bbox} (right). The result is usable, albeit presenting some ambiguity to subsequent algorithms. In the worse case such as case 2 where the time difference  between the two sensors $\Delta t_2$ is above the time tolerance, the 3D and 2D bounding boxes do not overlap (not shown), potentially hurting the path planning result.

In practice, IoU (intersection over union between two bounding boxes) can be used to determine the threshold of tolerable synchronization error. \Tbl{tab:tolerance} presents the relationship between velocity and tolerable synchronization error under different IoU requirements using parameters from our vehicles.

%%%%%%%%%%Good stuff below that we should save for a journal paper%%%%%%%%%%
%Assume that the projected 3D detection bounding box onto 2D image has the same size with the 2D detection bounding box, namely $B$. Both $B$ and the intersection between the 3D and 2D bounding boxes $I$ can be computed directly from the image. Given a vehicle with its length $l$ and its velocity $v$, the difference between the triggering times of the LiDAR and the camera can be estimated by $\Delta t=\frac{B-I}{B}\frac{l}{v}$. If the two boxes are matched, the IoU should be greater than a threshold $\theta _{IoU}$, i.e. ${\rm IoU}=\frac{I}{2B-I}>\theta _{IoU}$, so we have $\Delta t<\frac{1-\theta _{IoU}}{1+\theta _{IoU}}\frac{l}{v}$. To guarantee accurate matching, the synchronization error should not be greater than $\Delta t$.
%%%%%%%%%%%%%%%%%%%%%%%%%%%%%%%%%%

\begin{table}[t]
\caption{Tolerable synchronization errors for object detection under different velocities (assuming vehicle length $= 4.09$ m).}
\centering
\begin{threeparttable}
\begin{tabular}{c|ccccc}
\hline
\multicolumn{1}{c}{} & {Velocity (m/s)} & 5 & 10 & 20 & 40    \\ \hline
$\theta _{IoU}=0.5$ & Tolerance (ms) & 273 & 136 & 68 & 34      \\ 
$\theta _{IoU}=0.0$ & Tolerance (ms) & 818 & 409 & 205 & 102 \\ \hline
\end{tabular}
%\begin{tablenotes}
%\footnotesize
%\item $^*$Vehicle length $l=4.09m$.
%\end{tablenotes}
\end{threeparttable}
\label{tab:tolerance}
\end{table}

\paragraph{Localization} Ego-pose estimation in a reference frame, namely localization, is a fundamental task in autonomous machines. 
Modern autonomous machines usually adopt multi-sensor fusion techniques to achieve accurate ego-pose estimations. For instance, visual-inertial odometry (VIO), which fuses camera and IMU sensors, has been widely used in drones and autonomous vehicles \cite{qin2018vins}.

\begin{figure}[t]
%\vspace{-10pt}
\centering
\subfloat[Translation.]
{
  \includegraphics[trim=1 1 1 1, clip, width=0.5\columnwidth]{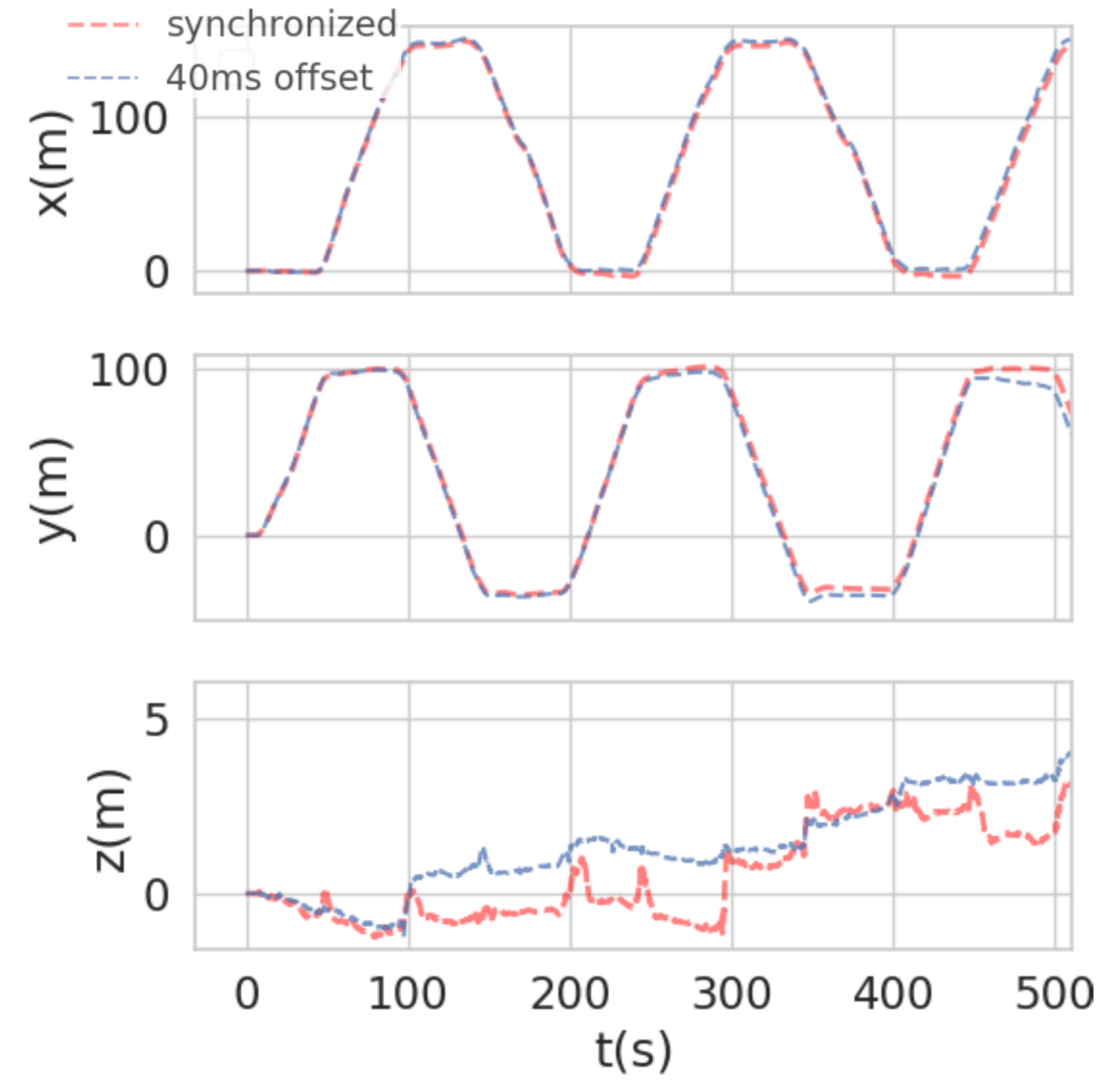}
  \label{fig:traj}
}
\subfloat[Rotation.]
{
  \includegraphics[trim=0 0 0 0, clip, width=0.5\columnwidth]{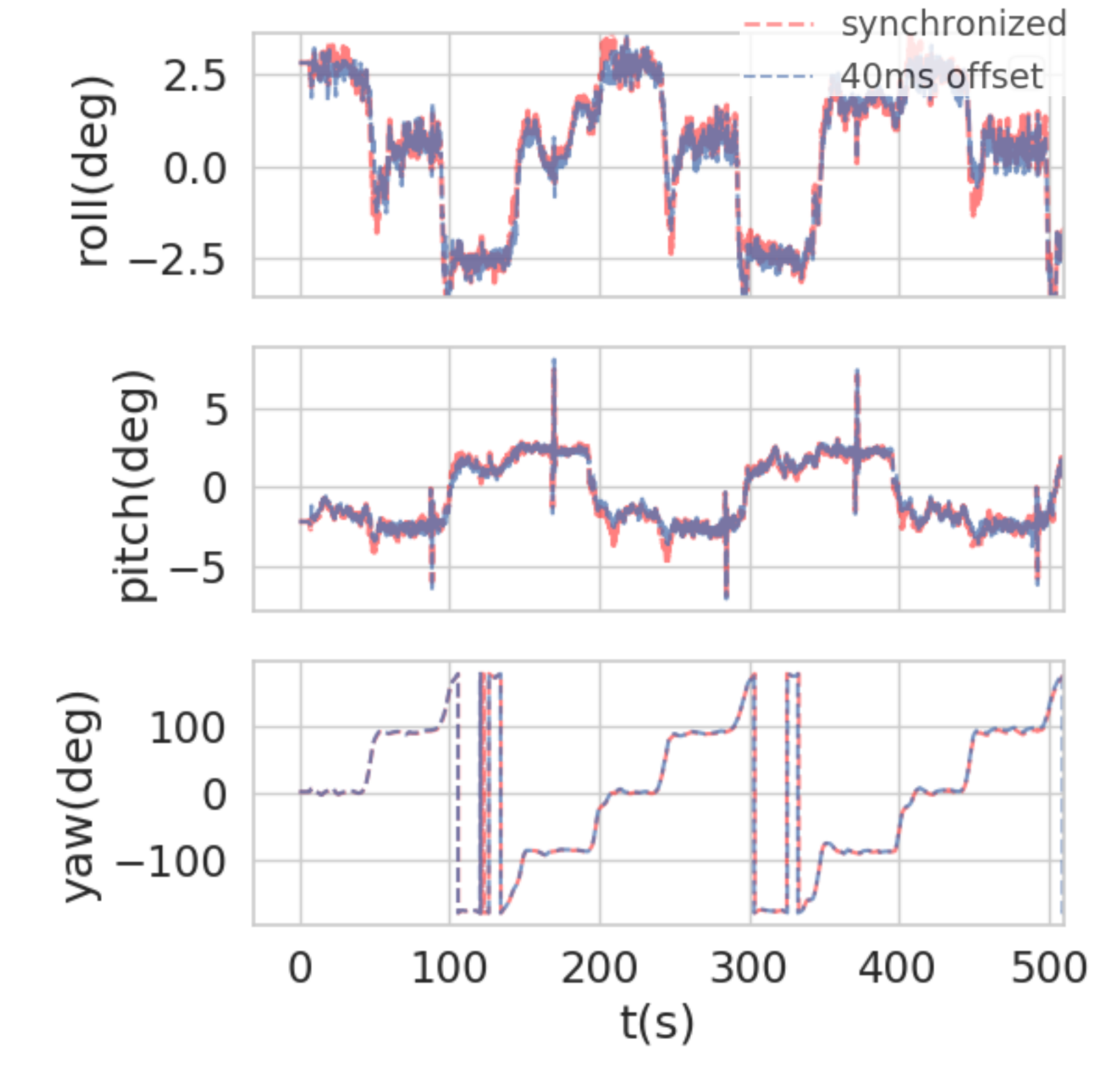}
  \label{fig:rotation}
}
\caption{VIO poses' discrepancies in translation ($x$, $y$ and $z$ axis) and rotation ($roll$, $pitch$ and $yaw$ axis) between using synchronized and mis-synchronized camera and IMU data.}
\label{fig:sync_unsync}
\vspace{-5mm}
\end{figure}

Mis-synchronized sensor data contributes temporal noises to the sensor fusion algorithm and leads to estimation errors in both translational and rotational poses. In \Fig{fig:sync_unsync}, we present data captured in real-world deployments and compare a VIO algorithm's performance between using synchronized and mis-synchronized data. When cameras and IMUs are off-sync by 40 ms, the translational error could be 10 m (Fig.~\ref{fig:traj}) and the rotational error could be as much as 3 degrees (Fig.~\ref{fig:rotation}). 
%\fixme{I didn't understand the figures.} \fixme{$>>yubo$ Have updated the caption of Fig. \ref{fig:sync_unsync}}

\subsection{The Need for Inter-Machine Synchronization}
\label{sec:synch:ext}

Inter-machine synchronization is critical when an autonomous machine interacts with other machines, in the case of multi-robot collaboration, or a road side unit (RSU), in the case of infrastructure-vehicle cooperative autonomous driving. Let us use object speed estimation from RSUs, which is a typical scenario in infrastructure-vehicle cooperative autonomous driving, as an example to demonstrate the impact of inter-machine synchronization.

An object's speed can be estimated by two consecutive observations from the RSUs. Theoretically, with a position of $p_1$ at timestamp $t_1$ reported from one RSU and a position of $p_2$ at timestamp $t_2$ reported from another RSU, the object speed is estimated by $s=\frac{|p_2-p_1|}{t_2-t_1}$. Now let's suppose that the vehicle's timer is well-synchronized with the first RSU, but is mis-synchronized with the second RSU by $\Delta t$. In this case, the timestamp of its observation of $p_1$ is exactly $t_1$, but the timestamp of the second observation of $p_2$ should have been $t_2 +\Delta t$. As a result, the estimated object speed should have been $s'=\frac{|p_2-p_1|}{t_2+\Delta t-t_1}$.

%%%%%%%%%%Good stuff below that we should save for a journal paper%%%%%%%%%%
%\begin{figure}[t]
%\centering
%\includegraphics[width=1\columnwidth]{figs/speederror.pdf}
%\caption{\fixme{can we generate a figure where x-axis is $\Delta t$, and y-axis is the error in speed estimation? Assume some values for $d_1$ and $d_2$.}}
%\label{fig:speederror}
%\vspace{-5mm}
%\end{figure}
%%%%%%%%%%%%%%%%%%%%%%%%%%%%%%%%%%

We use a real case to show how object speed estimation is affected by misaligned inter-machine synchronization: under a $t_1 = 6317$ ms, $t_2 = 6818$ ms, $\Delta t = 849$ ms, $p_1 = (180.388,463.93)$, and $p_2 = (177.235,463.749)$, the accurate and the estimated speeds are $s=6.30$ m/s and $s'=2.34$ m/s, respectively, leading to an error of $|s-s'|=3.96$ m/s. 

\section{Four Principles of Sensor Synchronization}
\label{sec:inext}

Recall that sensor synchronization aims to ensure that sensor samples that have the same timestamp correspond to the same event. This in turn translates to two requirements~\cite{yu2020building}:

\begin{enumerate}
\setlength{\itemindent}{-10pt}
	\item different sensors are triggered simultaneously, and 
	\item each sensor sample is associated with a precise timestamp.
\end{enumerate}

Meeting these two requirements is difficult, because modern autonomous machines are equipped with a wide range of different sensors that differ in their triggering mechanisms and time-stamping mechanisms. \Tbl{tab:sensor} summarizes the characteristics of common sensors used in today's autonomous machines. We now explain the principles behind our synchronization system, which addresses the sensor diversity to meet the two requirements above. We focus on the design principles here while leaving the system implementation to the next section.

\vspace{8pt}
\noindent\underline{\textbf{Principle 1}}: \textit{Trigger externally-triggered sensors in hardware.}

Camera and IMU are sensors that must be triggered by an external signal. The simplest way of triggering the camera and the IMU is from the software running on host machine. This approach, however, makes it hard to trigger the camera and IMU simultaneously, because the software stacks (e.g., drivers) of the camera and the IMU are different, and thus introducing different latencies from the triggering signal is issued from the host to when it arrives at the two sensors. Therefore, our system triggers the cameras and the IMUs through a custom hardware, which allows the triggering signal to be delivered to the sensors in a few clock cycles with little variation.

\vspace{8pt}
\noindent\underline{\textbf{Principle 2}}: \textit{Use a common, shared timer within the machine to simultaneously initiate the internally-triggered sensors and send the triggering signals to externally-triggered sensors.}

Not only must the camera and the IMU be triggered simultaneously with each other, they must also be triggered simultaneously with the LiDAR and Radar as well. LiDAR and Radar are internally-triggered; they, once received the initial ``start'' signal, are able to periodically output samples by themselves without the intervention of the host.

To synchronize these internally-triggered sensors with the externally-triggered sensors, we would synchronize the initial signals sent to the internally-triggered sensors with the triggering signals sent to the externally-triggered sensors. This synchronization can be done by referring to a common time source (timer) in an autonomous machine. As a perhaps obvious note, the triggering frequency of internally-triggered sensors must be divisible by that of the externally-triggered sensors, or vice versa.

\vspace{8pt}
\noindent\underline{\textbf{Principle 3}}: \textit{Synchronize the machine timer with the global atomic clock from the GPS to enable external synchronization.}

If only intra-machine synchronization is required, the timer in an autonomous machine could be inaccurate (i.e., different from the global time), as long as all the sensors share the same timer. For inter-machine synchronization, however, if different robots or autonomous vehicles use their own time sources that are unknown to each other, it would be very difficult to align sensor samples across machines on the same time scale.

Therefore, different autonomous machines must synchronize their time sources to a global, high-precision, atomic clock time source, which is usually chosen to be the time data transmitted through GPS satellites.

Traditionally, the time data from GPS is obtained through a GPS receiver and processed by the the Network Time Protocol (NTP) \cite{mills1991internet}, which has the advantage of being a software-only protocol natively supported in Linux. Our first generation of product uses NTP. However, the software variation of the NTP stack introduces as much as 20 ms time variation across different sensors, while we have a requirement of variation of less than 1 ms. Our current system thus uses the IEEE 1588 Precision Time Protocol (PTP)~\cite{correll2005design}, which reduces the time variation down to the 100 $\mu$s range, across multiple autonomous vehicles. Note that IEEE 802.1AS defines a generalized form of IEEE 1588, called gPTP, which is also widely used in autonomous machines.

\begin{table}[t]
\caption{Characteristic of different sensors.}
\resizebox{\columnwidth}{!}{
\renewcommand*{\tabcolsep}{3pt}
\begin{tabular}{ccccc}
\toprule[0.15em]
\multicolumn{1}{l}{} & \textbf{Camera} & \textbf{IMU} & \textbf{LiDAR} & \textbf{Radar}    \\
\midrule[0.05em]
Triggering mechanism    & External      & External    & Internal & Internal \\ 
Time-stamping mechanism & External            & External          & Internal      & External       \\
Sensor Interface            & MIPI          & Serial Port & Ethernet & CAN      \\
\bottomrule[0.15em]
\end{tabular}
}
\label{tab:sensor}
\vspace{-5mm}
\end{table}

\vspace{8pt}
\noindent\underline{\textbf{Principle 4}}: \textit{Obtain the timestamp for a sensor sample as close to the sensor source as possible.}

The three design principles above ensure that sensors, both within an autonomous machine and across autonomous machines, are triggered simultaneously, meeting the first requirement. Equally important is the second requirement, which is to associate each sensor sample with a precise timestamp.

Today's off-the-shell LiDARs, internally, can associate each point cloud frame with a precise GPS timestamp as LiDAR firmware implements the PTP --- an ideal scenario.

Sensor samples from the cameras, IMUs, and Radars must be externally time-stamped. The simplest approach would be to obtain the timestamp when the sensor sample arrives at the application running on the host. This approach, again, introduces variable latency throughout the software and hardware stack as the localization where the timestamp is obtained is far away from where the data is generated, i.e., the sensor source.

Instead, we obtain the timestamps for camera, IMU, and Radar at their sensor interface, i.e., where the sensor data is received by the Systems-on-a-chip (SoC) from the off-chip sensor. Note that the time when a sensor sample arrives at the sensor interface is still different from when it is triggered. However, the time difference is usually much shorter and more deterministic, and thus can be compensated in software later.

\section{FPGA-Based Synchronization System}
\label{sec:hardware}

Practicing the four design principles requires a collaboration between software and hardware. To that end, we choose FPGA-based SoCs to implement our synchronization system.
% We first describe our system implementation details with a first-order evaluation, followed by a discussion of how we see our implementation can be readily adapted in the future for sensor synchronization for space robotics.

\begin{figure}[t]
\centering
\includegraphics[width=0.85\columnwidth]{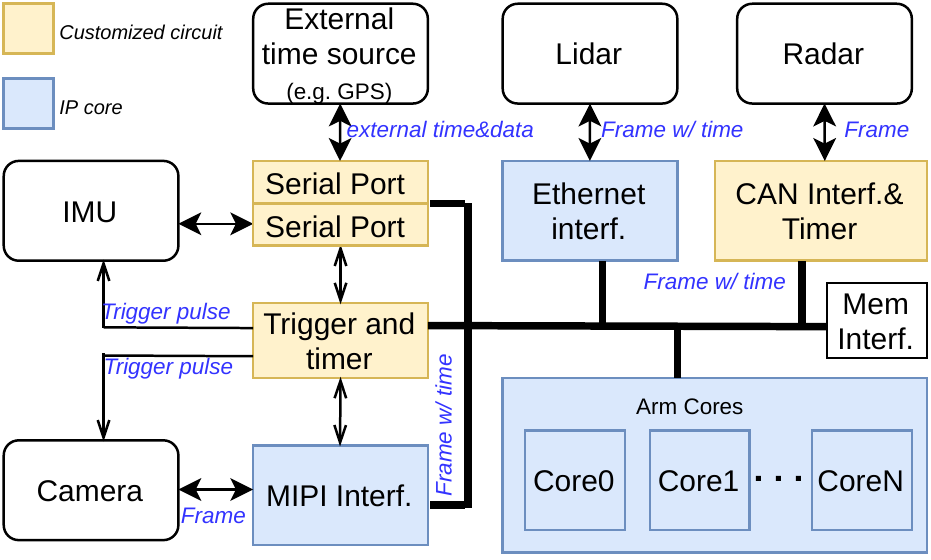}
\caption{FPGA-based synchronization system architecture.}
\label{fig:sync_sys}
\vspace{-5mm}
\end{figure}

\paragraph{System Implementation} Our synchronization system uses a Xilinx Zynq board, as shown in \Fig{fig:sync_sys}. Zynq SoCs contain both a general-purpose Arm CPU cluster and an FPGA fabric. Critically, these SoC boards are equipped with rich I/O interfaces required by different sensors (\Tbl{tab:sensor}).

The serial port interface accepts external global time data (e.g., GPS), which is distributed to the hardware timer and the Arm cores. %\fixme{The Arm CPUs communicate synchronization protocols with other clients.} 
The Trigger and Timer unit periodically generates trigger pulse to trigger the cameras and IMUs. The timestamp of an IMU sample, a camera sample, and a Radar sample is obtained and packed at the Serial Port interface, MIPI interface, and the CAN interface, respectively. LiDAR samples, along with the timestamps, are transmitted through the Ethernet interface. The Ethernet and MIPI interface implementations directly use the Xilinx IP cores, whereas the GPS, IMU and CAN interfaces, and the triggering circuits are customized in the FPGA fabric. The Arm cores are for the synchronization protocol, configurations, and user application interfaces.

\paragraph{Results} The customized circuits are lightweight, consuming 6.9K LUTs and 7.1K Regs. In terms of accuracy, Camera and IMU that use hardware triggering circuits achieve cycle-level precision on time-stamping. By using PTP protocol, LiDAR frames only exhibit variations within 100 $\mu$s.

\section{Related Work}

\paragraph{Synchronization on Earth} Sensor synchronization in robotics is vastly under-studied and public information is scarce. Most prior work in the literature either assumes that the sensors are perfectly synchronized, which is unrealistic in real-world scenario, or targets a specific application with a specific sensor setup. For instance, Nikolic et al.~\cite{nikolic2014synchronized} design an effective intra-machine synchronization system for cameras and IMUs and demonstrate its effectiveness in real-time SLAM tasks in mobile robots and Micro Aerial Vehicles. Similarly, Yu et al.~\cite{yu2020building} propose a system that synchronizes cameras and IMUs within autonomous machines. Their system practices our Design Principle 4, i.e., obtaining sensor timestamps closer to the sensor source. They do not, however, implement the other design principles and, thus, do not fully support intra-machine and inter-machine synchronization.

\paragraph{Supporting Sensor Synchronization in Space}
We expect that the demand for space exploration robots will drastically increase in the near future \cite{li2019enabling}, and thus time synchronization for space exploration robots is required. For intra-machine synchronization, the time source distribution mechanism stays the same. For inter-machine synchronization, the Proximity-1 protocol is used for orbiters and rovers to communicate with each other \cite{mills2017computer}. A typical spacecraft consists of a mission elapsed time counter called SCLK, which is calibrated with UTC (by control stations on Earth) and can be used by robots in space as the global time source. Thus, our current system design can be adapted to support inter-machine synchronization in space by simply replacing the GPS receiver with a SCLK signal receiver.

\section{Summary and Discussion}

This paper summarizes our past few years' experiences on developing time synchronization systems in our commercial robotic and autonomous driving products. Specifically, different robotic products have different intra-machine and inter-machine synchronization requirements, as they are equipped with different sensors and utilize different global time sources. To address these problems, we have developed a general hardware synchronization system for robotic applications. We have verified that our proposed design delivers high synchronization accuracy (in the $\mu$s range) while maintaining minimum energy and resource consumption.

\bibliographystyle{IEEEtran}
\bibliography{ref}
\end{document}